%% file: KrauseFischerIgelPopCDICLR-arXiv.tex
\documentclass[letterpaper]{article} 
\usepackage{iclr2016_conference,times}
\usepackage[pass]{geometry}
\pdfoutput=1 
\usepackage{hyperref}
\usepackage{url}

\usepackage{subfigure}
\usepackage{graphicx}
\usepackage{epstopdf}
\usepackage{tikz}
\usepackage{multirow}

\usepackage{algorithmic}
 \usepackage[vlined,linesnumbered,ruled,titlenotnumbered]{algorithm2e}
\usepackage{amsmath}
\usepackage{amssymb}
\usepackage{amsthm}
\usepackage{bbm}

\def\vec#1{\mathchoice{\mbox{\boldmath$\displaystyle#1$}}
{\mbox{\boldmath$\textstyle#1$}}
{\mbox{\boldmath$\scriptstyle#1$}}
{\mbox{\boldmath$\scriptscriptstyle#1$}}}

\newcommand{\Energy}[0]{\ensuremath{\mathcal{E}}}
\newcommand{\h}{\vec h}
\newcommand{\vis}{\vec v}

\newcommand{\BaS}{Bars\,\&\,Stripes}

\usepackage{color}

\newcommand{\cirm}[1]{}
\newcommand{\okrm}[1]{}
\newcommand{\afrm}[1]{}
\newcommand{\afm}[1]{}
\newcommand{\cim}[1]{}

\title{Population-Contrastive-Divergence:\\Does Consistency help with RBM training?}
\author{%
Oswin Krause\\
Department of Computer Science\\
University of Copenhagen, Denmark\\
\texttt{oswin.krause@di.ku.dk}
\And
Asja Fischer\\
Montreal Institute for Learning Algorithms\\
Universit{\'e} de Montr{\'e}al, Canada\\
\texttt{fischer@iro.umontreal.ca}
\And Christian Igel\\
Department of Computer Science\\
University of Copenhagen, Denmark\\
\texttt{igel@di.ku.dk}
}

\begin{document}

\maketitle 

\begin{abstract}
Estimating the log-likelihood gradient with respect to the parameters 
of a Restricted Boltzmann Machine (RBM) typically requires sampling 
using Markov Chain Monte Carlo (MCMC) techniques.  To save computation 
time, the Markov chains are only run for a small number of steps, which 
leads to a biased estimate.  This bias can cause RBM training algorithms 
such as Contrastive Divergence (CD) learning to deteriorate.  We adopt the 
idea behind Population Monte Carlo (PMC) methods to devise a new RBM training 
algorithm termed Population-Contrastive-Divergence (pop-CD).  Compared to CD, 
it leads to a consistent estimate and may have a significantly lower bias.
Its computational overhead is negligible compared to CD but
the variance of the gradient estimate increases. We experimentally 
show that pop-CD can significantly outperform CD. In many cases, we observed
a smaller bias and achieved higher log-likelihood values.
However, when the RBM distribution has many hidden neurons, 
the consistent estimate of pop-CD may still have a considerable bias and
the variance of the gradient estimate requires a smaller learning rate.
Thus, despite its superior theoretical properties, it is not advisable to use pop-CD 
in its current form on large problems.
\end{abstract}
\section{INTRODUCTION} 
Estimating the log-likelihood gradient with respect to the parameters of an 
undirected graphical model, such as a Restricted Boltzmann Machine
(RBM, \citealp{smolensky,HintonCD,fischer:13}), is a challenging task. As the analytic calculation of 
the gradient is infeasible, it is often approximated using Markov Chain Monte Carlo (MCMC)
techniques. Getting unbiased samples from the model distribution 
requires to run the Markov chain until convergence, but in
practice the chain is only iterated for a fixed amount of steps and the quality of the samples is often unknown.
\par Arguably, $k$-step Contrastive Divergence (CD-$k$, \citealp{HintonCD}) is 
the algorithm most often used for RBM training. In each training iteration of 
CD-$k$ a new Markov chain is initialized with a sample from the dataset, and
$k$ steps of {block-}Gibbs-sampling are performed.
In practice, $k$ is often set to one, which obviously results in a considerable bias of the gradient approximation.
Even though the bias is bounded \citep{FischerCDBias}, it was reported to cause divergence in the long 
run of the optimization \citep{FischerCDDivergence}, which is difficult to detect by heuristics \citep{schulz2010investigating}. 
Moreover, it has been shown that the gradient field of the CD
approximation does not belong to any objective function
\citep{sutskever2010convergence}.

\par After the introduction of CD, other sampling schemes were proposed
for the gradient approximation, the most prominent ones being 
Persistent Contrastive Divergence (PCD, \citealp{tieleman2008training}) 
and Parallel Tempering (PT, \citealp{SalakhutdinovTemperedTransitions, DesjardinsTempered}).
The former uses a persistent 
Monte Carlo Chain during training. While initially started from a
training example as in CD, the chain is not discarded after 
each gradient update but successively reused in following learning iterations. 
This is done with the hope that the chain stays close to the model distribution, 
while performing PCD-based gradient ascent. However, this requires a very small learning rate 
to guarantee that the model distribution changes slowly enough to compensate the small mixing rate of Gibbs-sampling. 
To solve this problem, PT was suggested for RBM training
\citep{SalakhutdinovTemperedTransitions, DesjardinsTempered}. This sampling method runs multiple 
tempered replica chains. A Metropolis-based swapping operator allows samples
to swap between the chains in order to achieve faster mixing. 
This is paid for by a higher computational cost.
While CD is inherently a parallel algorithm that can be applied simultaneously to the full data set, 
PT performs a step of all parallel Markov chains for every sample. This makes PT a serial 
algorithm that is difficult to transfer to GPUs, whereas implementing CD on a GPU 
is straightforward.
\par This paper introduces a different direction for RBM training that is based on the
Population Monte Carlo (PMC) method \citep{cappe2004population}. In PMC sampling, a set of 
samples is re-weighted using importance sampling after each 
application of the transition operator so that the weighted samples are unbiased. 
Furthermore, the weights can be used to re-sample the points. 
We will present a new algorithm combining importance sampling as in PMC
and  CD learning. It can be implemented as efficiently as CD without suffering
from a high bias {in RBMs with a small number of hidden neurons}.

\section{RBMS AND CONTRASTIVE DIVERGENCE} 
An RBM is an undirected graphical model with a bipartite structure. 
The standard binary RBM consists of $m$ visible variables 
$\vec V = (V_1,\dots V_m)$ taking states $\vis \in \{0,1\}^m$ and $n$ hidden variables 
$\vec H = (H_1,\dots H_n)$ taking states $\h \in \{0,1\}^n$.
The joint distribution is a Gibbs distribution 
$p(\vis,\h)=\frac{1}{Z} \tilde{p}(\vis,\h) = \frac{1}{Z} e^{-\Energy(\vis, \h)}$
with energy $\Energy(\vis,\h)=-\vis^T \vec W \h - \vis^T\vec b -\vec c^T\vec{h}$,
where $\vec W, \vec b, \vec c $ are the weight matrix 
and the visible and hidden bias vectors, respectively. The normalization constant  
$Z = \sum_{\vis}\sum_{\h} e^{-\Energy(\vis, \h)} $ (also referred to as partition
function) is typically unknown, because it is calculated by summing 
over all possible states of hidden and visible units, 
which is exponential in $\min\{n,m\}$.
\par In this paper, we focus on the problem of maximum log-likelihood fitting of the RBM distribution to a 
data set $S=\{\vec v_1,\dots \vec v_{\ell}\}, \vec v_i \in \{0,1\}^m$. The gradient of $ \log p(\vec x_i)$
w.r.t.~the model parameters $\vec \theta=(W,\vec b,\vec c)$ is given by
\begin{equation}\label{GradientRBM}
\frac {\partial \log p(\vec v_i)}{\partial \vec \theta} = 
	E_{p(\h| \vec v_i)}\left\{ \frac {\partial \Energy(\vec v_i,\h)}{\partial \vec \theta}  \right\}
	-E_{p(\vis, \h')}\left\{\frac {\partial \Energy(\vec \vis,\h')}{\partial \vec \theta} \right\} \enspace.
\end{equation}
While the first term of the derivative can be computed analytically, 
the calculation of the second term requires to sum over all $\vis \in \{0,1\}^m$, which
is intractable. Instead, samples are used to obtain an estimate of the expectation under the RBM distribution. 
The RBM training algorithm most often used in literature is CD, which approximates the expectation over 
$p(\vis,\h')$ by a sample gained after $k$ steps of Gibbs-sampling starting from sample $\vec v_i$. 

\subsection{Population Monte Carlo}
Being a Markov chain Monte Carlo (MCMC) method, Population Monte Carlo
(PMC, \citealp{cappe2004population}) aims at 
estimating expectations of the form
\begin{equation} \label{eq:expectation}
E_{p(\vec x)}\left\{f(\vec x) \right\}= \int p(\vec x) f(\vec x) \text{d}\vec x \enspace.
\end{equation}
When samples $\vec x_1,\dots \vec x_{N}$ from $p$ are available, this expectation can be approximated by
\begin{equation}\label{MCMCEstimate}
E_{p(\vec x)}\left\{f(\vec x) \right\}\approx 
\frac 1 N \sum_{i=1}^{N}f(\vec x_i)\enspace.
\end{equation}
Often it is not possible to generate samples from $p$ directly.
In this case, MCMC methods provide a way to sample from $p$
by running $k$ steps of a Markov chain that has $p$ as its stationary
distribution.
\par In practice, it is computationally too demanding to choose $k$ 
large enough to guarantee convergence. Consequently, the samples 
$\vec x_i$ generated by the Markov chain are not drawn from $p$ but from another
distribution which we will refer to as $q$. Thus, using the samples directly in \eqref{MCMCEstimate} leads to a 
biased estimate for \eqref{eq:expectation}. 
\par In this case, importance sampling, where $q$ is regarded  as proposal distribution, could allow
for getting an unbiased estimate of \eqref{eq:expectation}  based on samples from $q$:
\begin{equation}\label{ImportanceSampling}
E_{p(\vec x)}\left\{f(\vec x) \right\}= E_{q(\vec x)}\left\{\frac{p(\vec x)}{q(\vec x)}f(\vec x) \right\}
\end{equation}
\par However, the distribution $q$ that results from running a Markov chain for $k$-steps is usually 
unknown and therefore a weighting scheme as in equation \eqref{ImportanceSampling} 
can not be applied. Instead one can use any known conditional distribution $\kappa(\vec x'| \vec x) > 0$
as proposal distribution for importance sampling since
\begin{equation}\label{PopMCReweighting}
E_{p(\vec x)}\left\{f(\vec x) \right\}= E_{q(\vec x)}\left\{
	E_{\kappa(\vec x'| \vec x)}\left\{\frac{p(\vec x')}{\kappa(\vec x'|\vec x)}f(\vec x') \right\}
\right\}\enspace.
\end{equation}
This re-weighting scheme does not require the knowledge of the probabilities $q(\vec x)$  
but just the ability to sample from $q$.
\par Equation \eqref{PopMCReweighting} requires that we know
$p(\vec x)$, but in probabilistic modelling often 
only  the unnormalized probability $\tilde{p}(\vec x)=Z p(\vec x)$ is available. 
Thus, $Z$  needs to be estimated as well.
Since $E_{\kappa(\vec x'| \vec
  x)}\left\{\frac{\tilde{p}(\vec x')}{\kappa(\vec x'|\vec x)}\right\}
= Z$ for all $\vec x $,
we get a biased importance sampling estimate by
\begin{equation}\label{MCEstimator}
E_{p(\vec x)}\left\{f(\vec x) \right\} \approx \sum_{i=1}^N \frac  {\omega_i f(\vec x_i)} {\sum_{j=1}^N \omega_j}\enspace,
\end{equation}
where $\vec x'_i \sim q(\cdot) $, $\vec x_i \sim \kappa( \cdot | \vec x'_i)$, and $\omega_i = \tilde{p}(\vec x_i)/\kappa(\vec x_i| \vec x'_i)$. 
The estimate becomes unbiased in the limit of $N \rightarrow \infty$ and thus the estimator is consistent.
\par A PMC method makes use of equations \eqref{PopMCReweighting} and \eqref{MCEstimator} 
to create a Markov chain, where in step $t$  states 
$\vec y_i^{(t)}\sim \kappa(\cdot|\vec x_i^{(t-1)})$, $i=1,\dots, N$ are
sampled.
For the $N$ states importance weights $\omega_i$ are estimated
and the states are then re-sampled accordingly. 
The re-sampling procedure results in the samples $\vec x_1^{(t)}, \dots, \vec x_N^{(t)}$ which 
corresponds to the $t$-th state of the chain. 
The chain thus only depends on the choice of $\kappa(\cdot|\vec x)$ which should be 
chosen to have heavy tails as to reduce the variance of the estimate.

\section{POPULATION-CONTRASTIVE-DIVERGENCE} 

In this section, we introduce the main contribution of this paper,
the Population-Contrastive-Divergence (pop-CD) algorithm
for RBM training. This algorithm exploits
the features of CD and Population-MCMC by combining the efficient implementation of CD-$k$ 
with the low bias re-weighting scheme introduced by PMC. 
\par Let us rewrite the log-likelihood gradient in equation~\eqref{GradientRBM} 
by setting $f(\vis)=E_{p(\h''|\vis)}\left\{\frac {\partial \Energy(\vec \vis,\h'')}{\partial \vec \theta} \right\}$ 
and estimating $E_{p(\vis)}\left\{f(\vis) \right\}=E_{p(\vis, \h'')}\left\{\frac {\partial \Energy(\vec \vis,\h'')}{\partial \vec \theta} \right\}$
based on the re-weighting scheme of equation~\eqref{PopMCReweighting}. 
For this, we choose $q$ to be the distribution of the hidden variables generated by
$k-1$-steps of block-wise Gibbs-sampling when starting from the $i$-th training example $\vis_i$,
which we will call $q_i^{(k-1)}(\h')$ in the following.
The proposal distribution $\kappa$ is set to the conditional distribution of the visible 
variables  given the state of the hidden, i.e., to $p(\vis|\h')$. This yields
\begin{equation}
	\frac {\partial \log p(\vis_i)}{\partial \vec \theta} =  
	E_{p(\h| \vis_i)}\left\{ \frac {\partial \Energy(\vis_i,\h)}{\partial \vec \theta}  \right\}
	-E_{q_i^{(k-1)}(\h')}\left\{E_{p(\vis|\h')}\left\{\frac{p(\vis)}{p(\vis|\h')} 
	f(\vis) \right\}\right\} \label{PMCGradient} \enspace.
\end{equation}
In the case of $k=1$ we can write 
$q_i^{(0)}(\h')=  p(\h'|\vec v_i)$
and for $k\geq 2$ the distribution $q_i^{(k-1)}$ can be written as a recursive relation over the distribution of $q_i^{(k-2)}$:
\[
	q_i^{(k-1)}(\h') = \sum_{\h'' \in H}\sum_{\vis \in V}p(\h'|\vis)p(\vis|\h'')q_i^{(k-2)}(\h'') 
\]
Now, based on the considerations in the previous section, it follows from \eqref{PMCGradient}  
that we can estimate the second term of the gradient 
given the training set $S=\{\vec v_1,\dots \vec v_{\ell}\}$
by \eqref{MCEstimator},
where  $\h_i' \sim q_i^{(k-1)}(\cdot) $,
$\vis_i' \sim p(\cdot|\h_i')$ and $\omega_i = \tilde{p}(\vec v_i')/p(\vec v_i'| \vec h'_i)$.
This results in the following estimate of the log-likelihood gradient
\begin{equation}
\frac{1}{\ell} \sum_{i=1}^{\ell} f(\vis_i)
- \sum_{i=1}^{\ell} \frac  {\omega_i f(\vec v_i')} {\sum_{j=1}^{\ell} \omega_j} =
\frac{1}{\ell} \sum_{i=1}^{\ell} E_{p(\h| \vis_i)}\left\{\frac {\partial \Energy( \vis_i,\h)}{\partial \vec \theta}\right\} 
- \sum_{i=1}^{\ell} \frac  {\omega_i E_{p(\h| \vis_i')}\left\{
	\frac {\partial \Energy( \vis_i',\h)}{\partial \vec \theta}\right\} 
} {\sum_{j=1}^{\ell} \omega_j}\label{popCDEstimate}\enspace,
\end{equation}
which we will refer to as pop-CD-$k$.
Setting $\omega_i =1$ or $\ell=1$ leads to CD-$k$.

In an actual implementation, it is advisable to work with
$\log\omega$ for increased numerical stability.
Algorithm \ref{alg:pop-CD} in the Appendix describes pop-CD in 
pseudo code.

\par In the following, we will analyse the properties of pop-CD-$k$ in more detail. We will analyse
the runtime costs as well as the bias of the method, take a closer look at the behavior of the weights, and
discuss alternative weighting schemes.

\subsection{Runtime}
Equation~\eqref{popCDEstimate} does not produce
a noteworthy computational overhead compared to CD,  because the only entities
to compute additionally are the weights $\omega_i$. The weights in turn are inexpensive to compute
as the distribution $p(\vis|\h^{(k-1)})$ is already computed  (see Algorithm \ref{alg:pop-CD}, line 
\ref{PVHAlgorithm}) and calculating $p(\vis^{(k)})$ can reuse
$\vec W\vis^{(k)}$, which has already been computed for determining
$p({H_i} = 1 \,|\, \vis^{(k)})$ for the gradient update.
Thus the cost to compute $\omega_i$ is negligible compared to the sampling cost.

\subsection{Bias}
It is known that equation~\eqref{MCEstimator} is biased for two reasons. The first reason is that
the same samples are used to estimate the gradient as for the estimation of $Z$. The second one is
that, even assuming an unbiased estimator for $Z$, this does not lead to an unbiased estimate of $1/Z$.
It is easy to remove the bias from using the same samples by rewriting 
\eqref{MCEstimator} as
\begin{equation}\label{UnbiasedMCEstimator}
	E_{p(\vec x)}\left\{f(\vec x) \right\} \approx \frac {N-1} N \sum_{i=1}^N \frac  {\omega_i f(\vec x_i)} {\sum_{j=1,j\neq i}^N \omega_j}\enspace,
\end{equation}
that is, by simply removing the $i$-th sample from the estimator of $Z$ when estimating $p(\vec x_i)$.
This yields an estimator which converges in expectation over all possible sets of $N$ samples to
\begin{equation*}
	E\left\{\frac {N-1} N \sum_{i=1}^N \frac  {\omega_i f(\vec x_i)} {\sum_{j=1,j\neq i}^N \omega_j}\right\}=\\
	Z E\left\{{\frac {N-1}{\sum_{j=1}^{N-1} \omega_j}}\right\}E_{p(\vec x)}\left\{f(\vec x)\right\}\enspace,
\end{equation*}
which is still biased in the regime of finite sample size since $E\left\{\frac {N-1}{\sum_{j=1}^{N-1} \omega_j}\right\} \neq \frac 1 Z$.
Applying this result to the problem of estimating the log-likelihood 
gradient \eqref{GradientRBM}, the pop-CD gradient converges to
\begin{equation*}
	E_{p(\h| x_i)}\left\{ \frac {\partial \Energy(\vec x_i,\h)}{\partial \vec \theta}  \right\} \\
	- {Z E\left\{\frac {N-1}{\sum_{j=1}^{N-1} \omega_j}\right\}}
	E_{p(\vis,\h)}\left\{ \frac {\partial \Energy(\vis,\h)}{\partial \vec \theta}\right\}\enspace.
\end{equation*}
Thus, using\eqref{UnbiasedMCEstimator}, only the length 
of the right term of the gradient will be biased and not its direction.

\par In practice, it is not advisable to employ \eqref{UnbiasedMCEstimator} instead of \eqref{MCEstimator} as usually only a small number of samples is used.
In this case, the estimate of $Z$ will have a high variance and the length of the gradient obtained by $\eqref{UnbiasedMCEstimator}$ can
vary by several orders of magnitude. This variance leads to
a small effective sample size in \eqref{MCEstimator}.

\subsection{Analysis of Weights}\label{Seq:AnalysWeights}
This section gives a deeper insight into the weights of the samples.
First, we rewrite the weight formula as
\begin{equation}
\omega= \frac{p(\vis)}{p(\vis|\h')}= p(\h') + \sum_{\h \neq \h'} p(\h) \frac{p(\vis|\h)}{p(\vis|\h')}\enspace.
\end{equation}
Now it becomes clear that the weight of a sample  $\vis \sim p(\vis|\h')$
is large if $\h'\sim q_i^{(k-1)}$ also has a high probability under
$p$ and that the weight is small otherwise.
Therefore, when $q_i^{(k-1)}$ puts more probability mass on samples less likely
under $p$, this is partly counterbalanced by the weighting scheme. The only case where
a low probability sample $\h'$ could lead to a large weight of a sample $\vis \sim p(\vis|\h')$ is
when $\frac{p(\vis|\h)}{p(\vis|\h')}$ is large for some $\h \neq \h'$. However, this would require 
a much smaller probability of the sample $\vis$ under $p(\vis|\h')$ than
under $p(\vis|\h)$, and, thus, the probability of drawing such a $\vis$ from  $p(\vis|\h')$ 
would be small.

\subsection{Alternative formulations}
In our approach, the weights $\omega_i=\tilde p(\vis_i')/p(\vis'|\h_i')$
solely depend on the current samples $\h'_i$. Therefore, only little
information about the distribution $q(\h')$
enters the learning process.
Now, one could ask the question whether it is beneficial to use the full sample $(\vis',\h')$
from the sampling distribution $q$ instead and defined in equation
\eqref{PMCGradient} the proposal distribution to be $\kappa(\vis,\h|\vis',\h')$.  
In this case, the weights become $\omega=\tilde p(\vis,\h)/\kappa(\vis,\h|\vis',\h')$. 
As the typical choice for $q$ is the block-wise Gibbs-sampler, 
we get $\kappa(\vis,\h|\vis',\h')=p(\vis|\h)p(\h|\vis')$ and 
$\omega=\tilde p(\h)/p(\h|\vis')$, which again depends on a single $\vis'$.
\par A different approach is to tie several samples of 
$q(\h')$ together in a mixture proposal distribution 
$\kappa(\vis|\h'_1,\dots,\h'_{\tau})=\frac 1 {\tau} \sum_i p(\vis|\h'_i)$. 
The advantage of this approach is that the 
proposal distribution takes distribution information of $\h'$ into
account. In fact,
$q(\vis|\h'_1,\dots,\h'_{\tau}) \overset{{\tau}\rightarrow
  \infty}{\longrightarrow} \sum_{\h\in H} q(\h) p(\vis|\h)$. The latter is a good proposal distribution
assuming $q(\h) \approx p(\h)$, however, the computational cost of computing the importance weights 
rises linearly with the number of samples in the mixture.

\section{EXPERIMENTS}
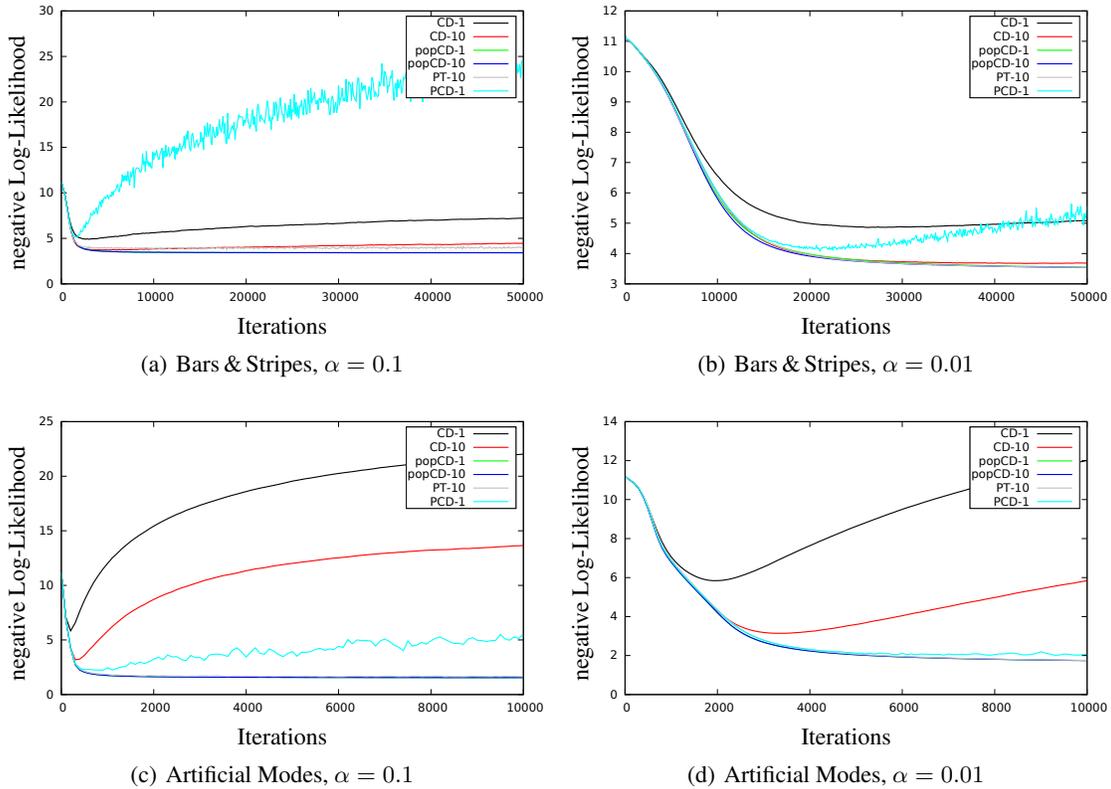
\begin{figure*}[t!]
	\hspace{-0.5cm}\subfigure[\BaS, $\alpha=0.1$]{
		\begin{tikzpicture}
			\draw (0, 0) node[inner sep=0] {\includegraphics[width=0.5\textwidth]{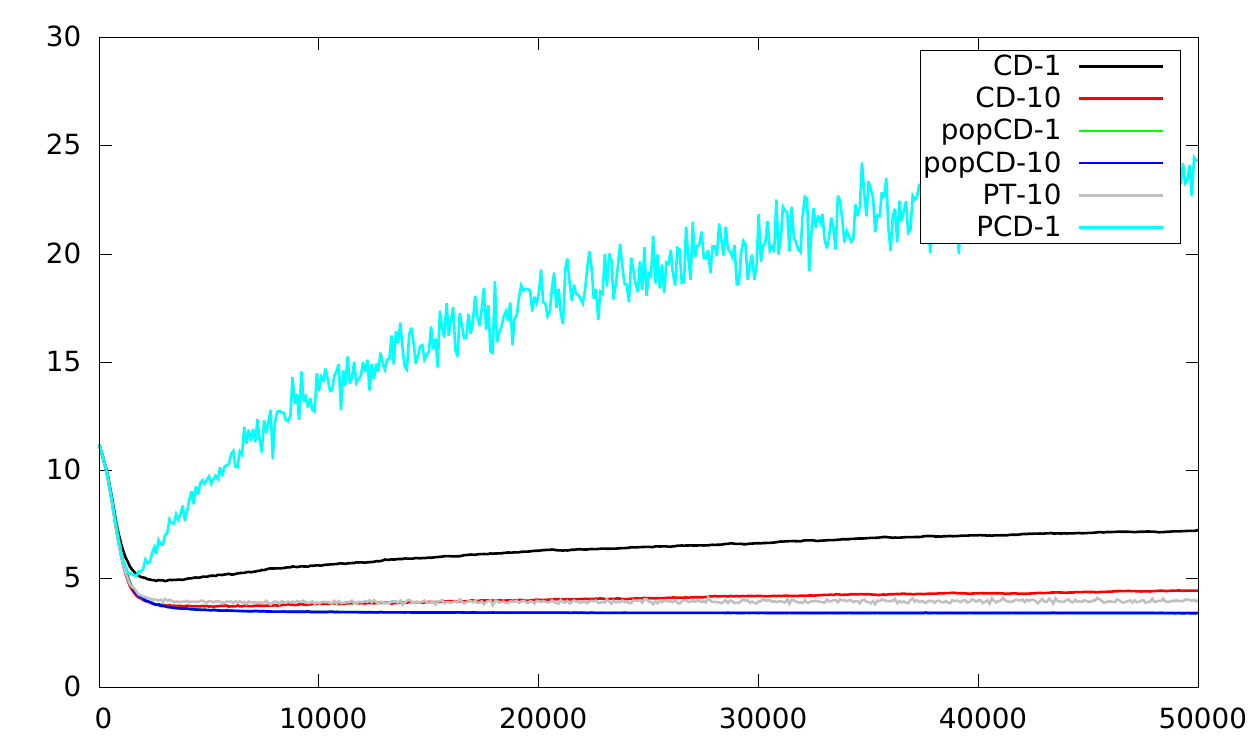}};
			\draw (-3.5, 0) node {\rotatebox{90}{\small negative Log-Likelihood}};
			\draw (0.0, -2.3) node {\small Iterations};
		\end{tikzpicture}
	}
	\subfigure[\BaS, $\alpha=0.01$]{
		\begin{tikzpicture}
			\draw (0, 0) node[inner sep=0] {\includegraphics[width=0.5\textwidth]{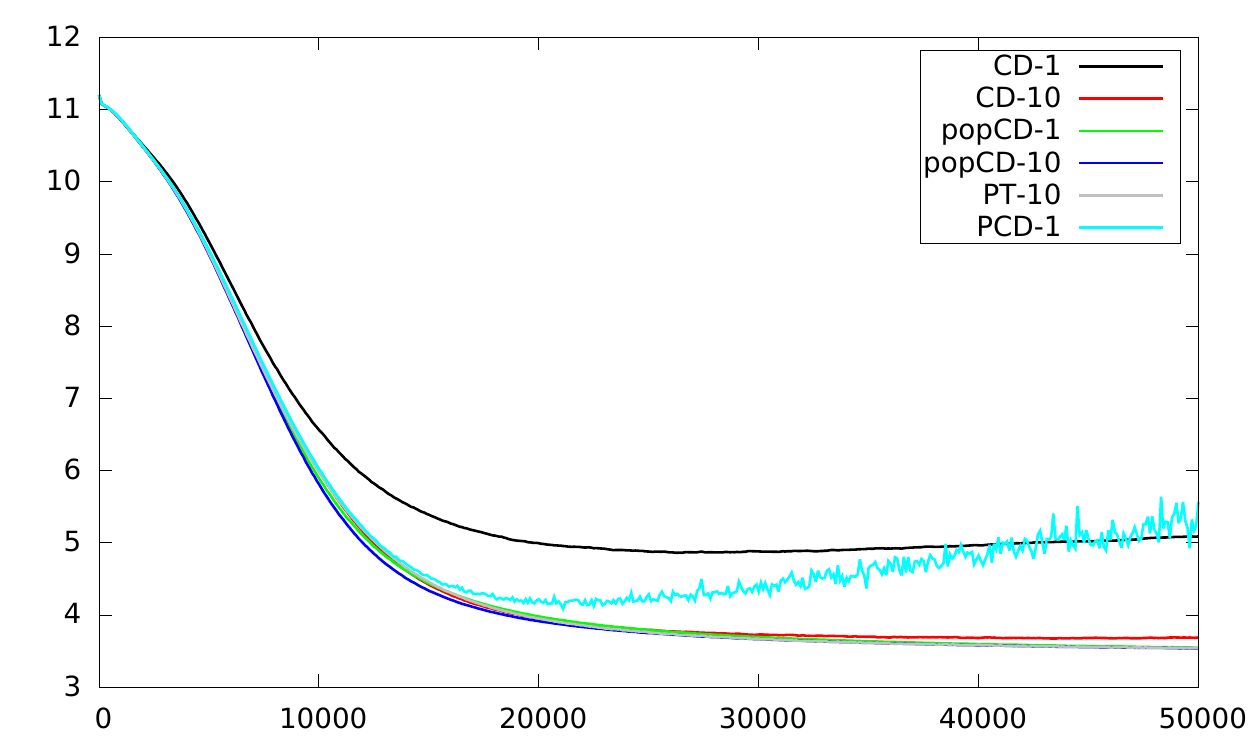}};
			\draw (-3.5, 0) node {\rotatebox{90}{\small negative Log-Likelihood}};
			\draw (0.0, -2.3) node {\small Iterations};
		\end{tikzpicture}	
	}\\
	\hspace*{-0.5cm}\subfigure[Artificial Modes, $\alpha=0.1$]{
		\begin{tikzpicture}
			\draw (0, 0) node[inner sep=0] {\includegraphics[width=0.5\textwidth]{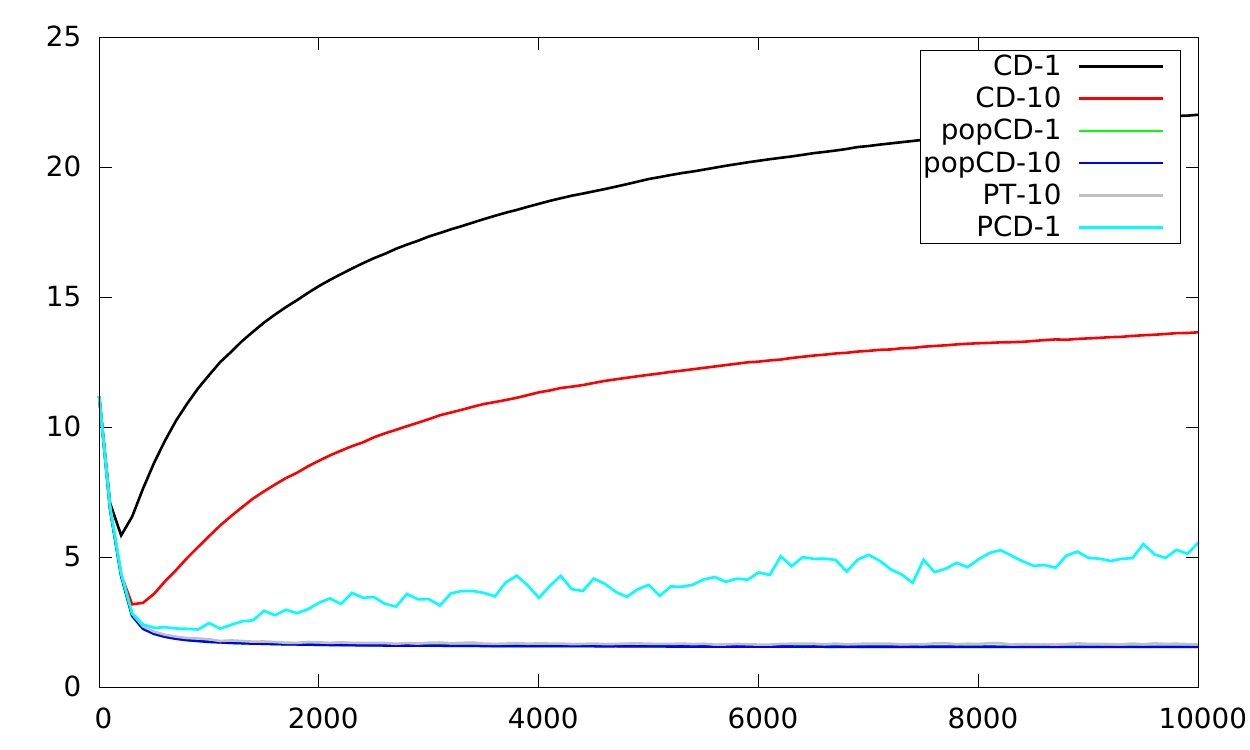}};
			\draw (-3.5, 0) node {\rotatebox{90}{\small negative Log-Likelihood}};
			\draw (0.0, -2.3) node {\small Iterations};
		\end{tikzpicture}
	}
	\subfigure[Artificial Modes, $\alpha=0.01$]{
		\begin{tikzpicture}
			\draw (0, 0) node[inner sep=0] {\includegraphics[width=0.5\textwidth]{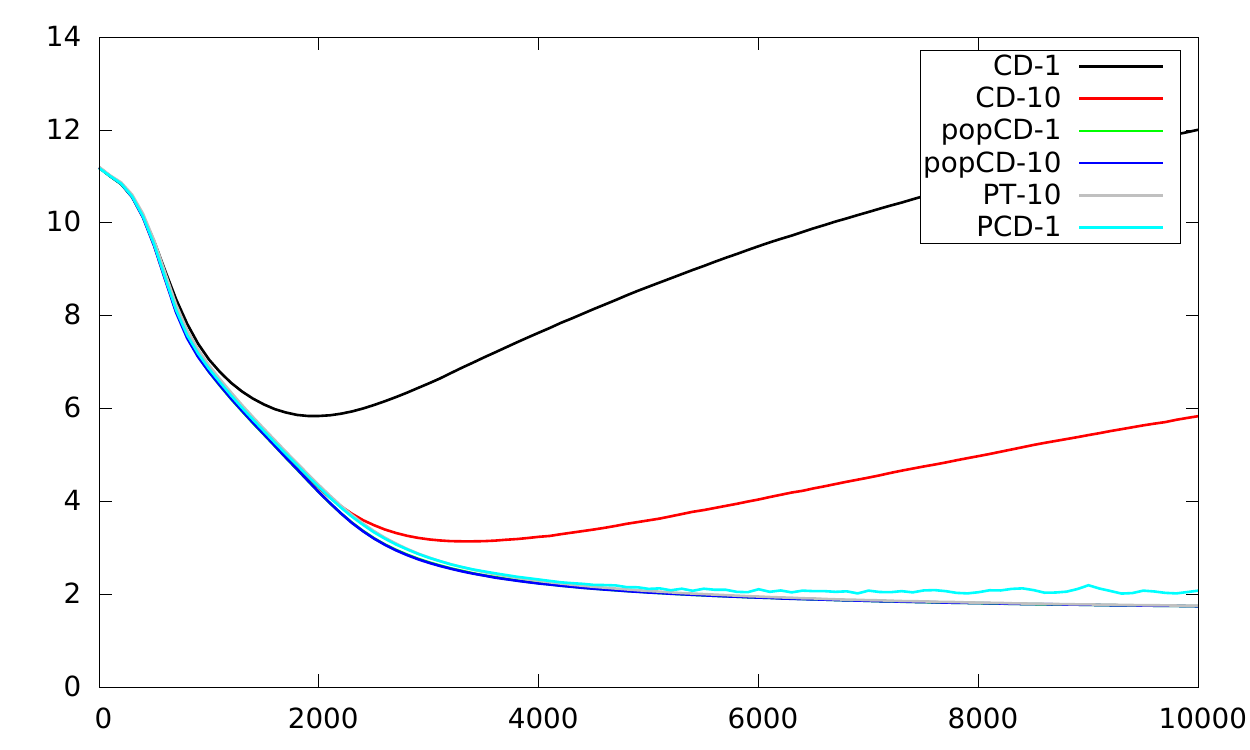}};
			\draw (-3.5, 0) node {\rotatebox{90}{\small negative Log-Likelihood}};
			\draw (0.0, -2.3) node {\small Iterations};
		\end{tikzpicture}	
	}\\
        \caption{Training curves for \BaS{} and Artificial Modes  with 16
          hidden neurons and different learning rates. The two
            pop-CD variants cannot be distinguished. Furthermore, the
            PT curve cannot be distinguished from pop-CD on Artificial Modes. }
	\label{SmallResults1}
\end{figure*}
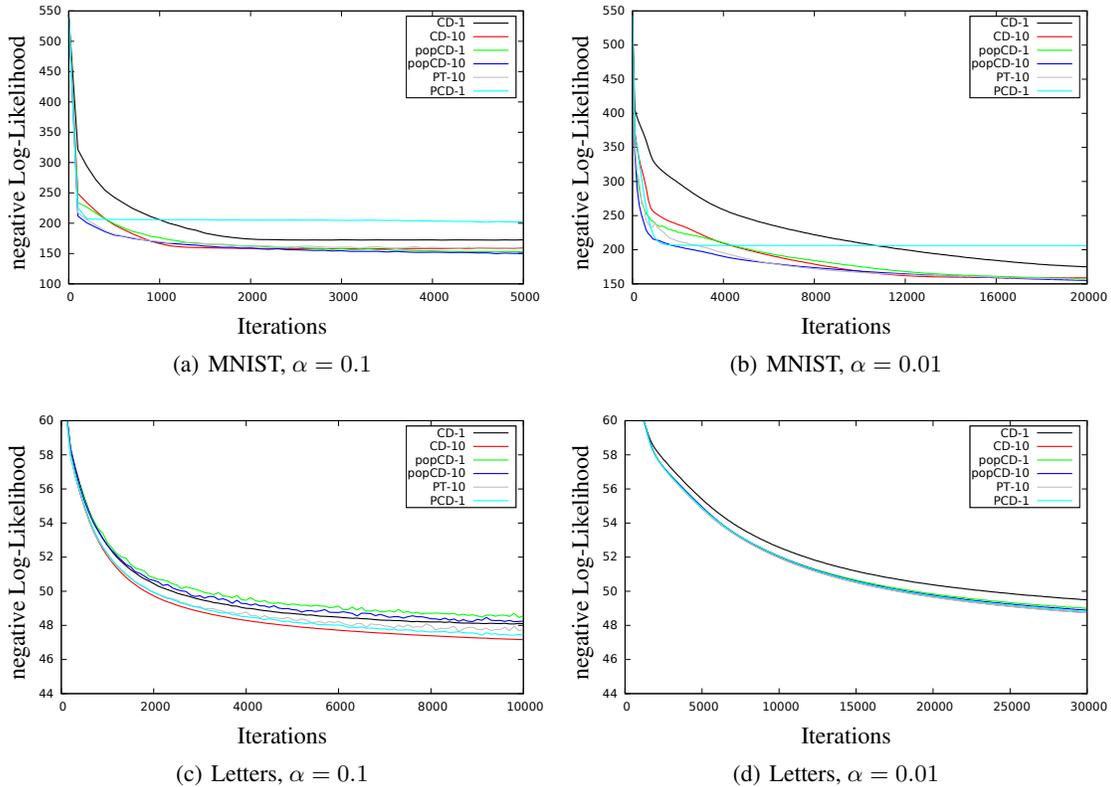
\begin{figure*}[t!]
	\hspace{-0.5cm}\subfigure[MNIST, $\alpha=0.1$]{\label{MNIST1601}
		\begin{tikzpicture}
			\draw (0, 0) node[inner sep=0] {\includegraphics[width=0.5\textwidth]{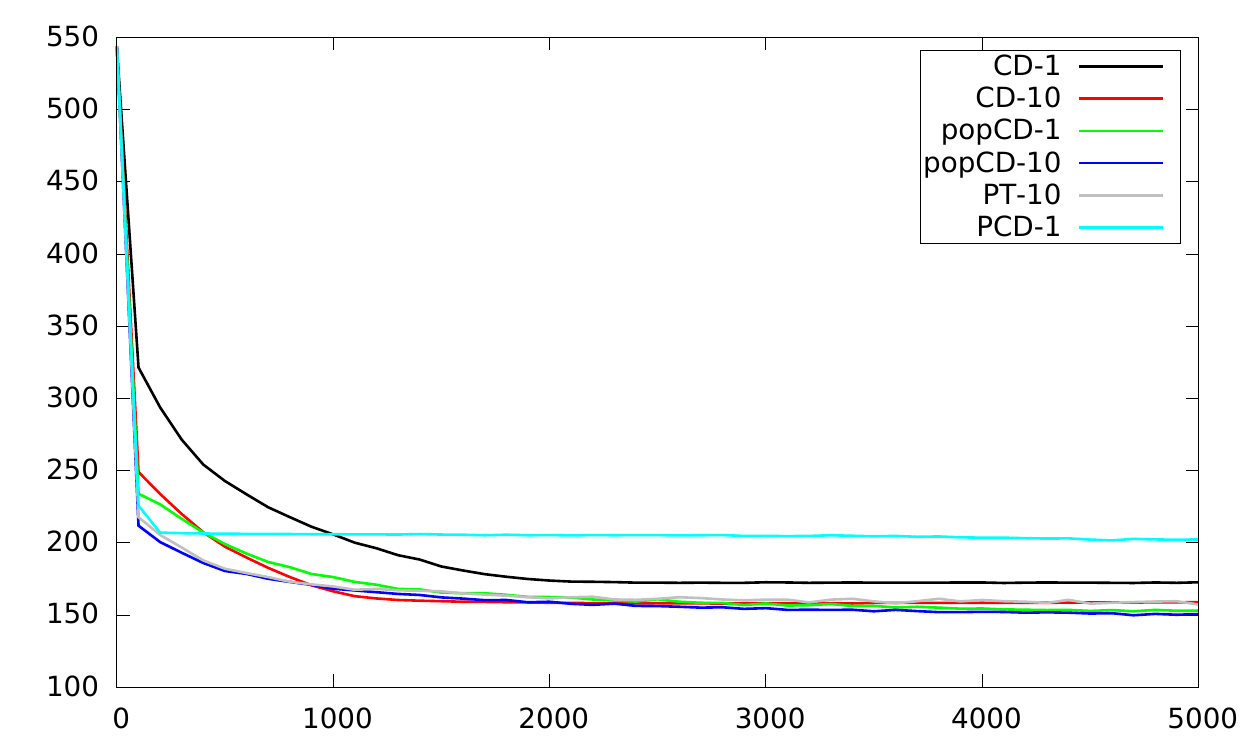}};
			\draw (-3.5, 0) node {\rotatebox{90}{\small negative Log-Likelihood}};
			\draw (0.0, -2.3) node {\small Iterations};
		\end{tikzpicture}
	}
	\subfigure[MNIST, $\alpha=0.01$]{
		\begin{tikzpicture}
			\draw (0, 0) node[inner sep=0] {\includegraphics[width=0.5\textwidth]{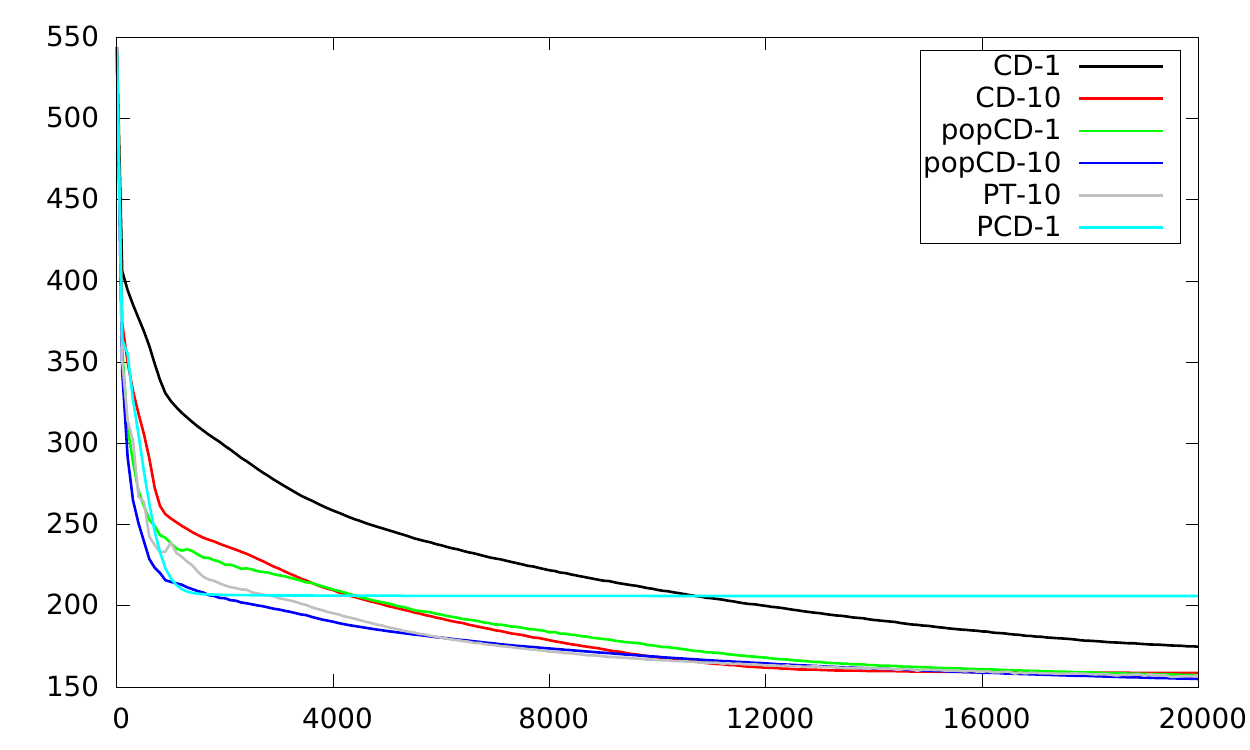}};
			\draw (-3.5, 0) node {\rotatebox{90}{\small negative Log-Likelihood}};
			\draw (0.0, -2.3) node {\small Iterations};
		\end{tikzpicture}	
	}\\
	\hspace*{-0.5cm}\subfigure[Letters, $\alpha=0.1$]{
		\begin{tikzpicture}
			\draw (0, 0) node[inner sep=0] {\includegraphics[width=0.5\textwidth]{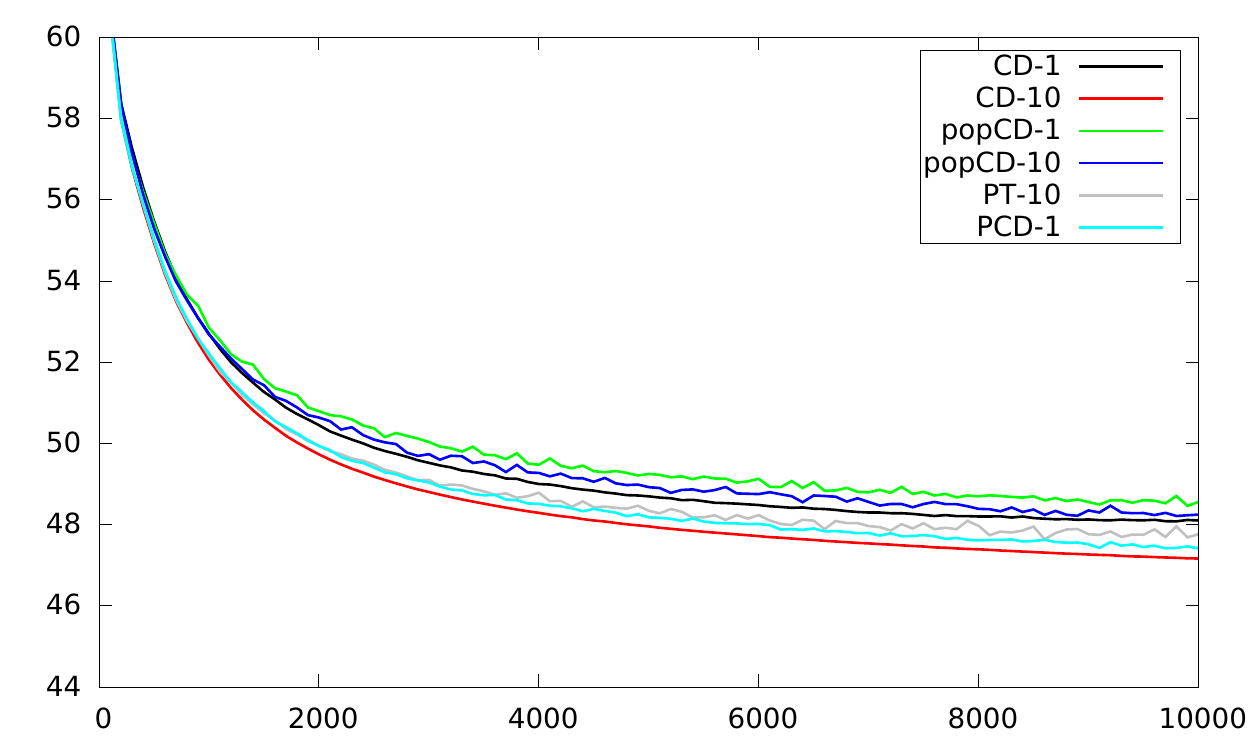}};
			\draw (-3.5, 0) node {\rotatebox{90}{\small negative Log-Likelihood}};
			\draw (0.0, -2.3) node {\small Iterations};
		\end{tikzpicture}
	}
	\subfigure[Letters, $\alpha=0.01$]{
		\begin{tikzpicture}
			\draw (0, 0) node[inner sep=0] {\includegraphics[width=0.5\textwidth]{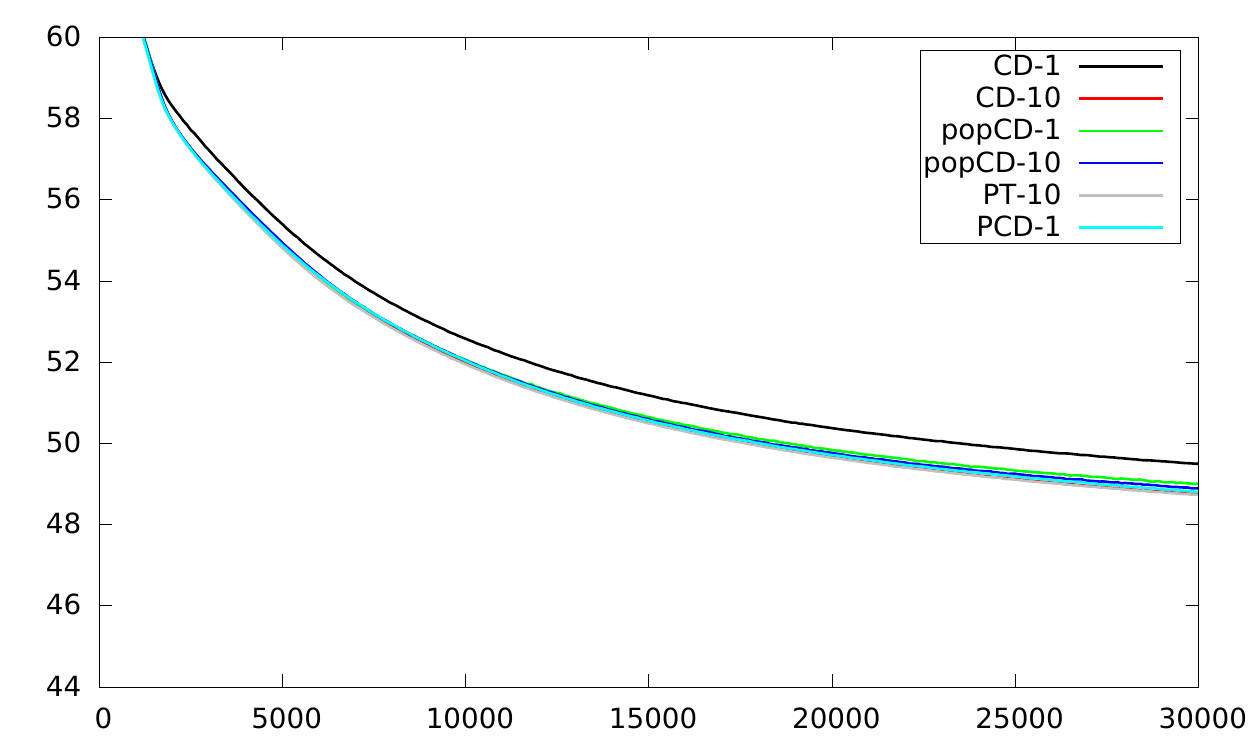}};
			\draw (-3.5, 0) node {\rotatebox{90}{\small negative Log-Likelihood}};
			\draw (0.0, -2.3) node {\small Iterations};
		\end{tikzpicture}	
	}\\
        \caption{Training curves for MNIST and Letters with 16 hidden neurons.}
	\label{SmallResults2}
\end{figure*}
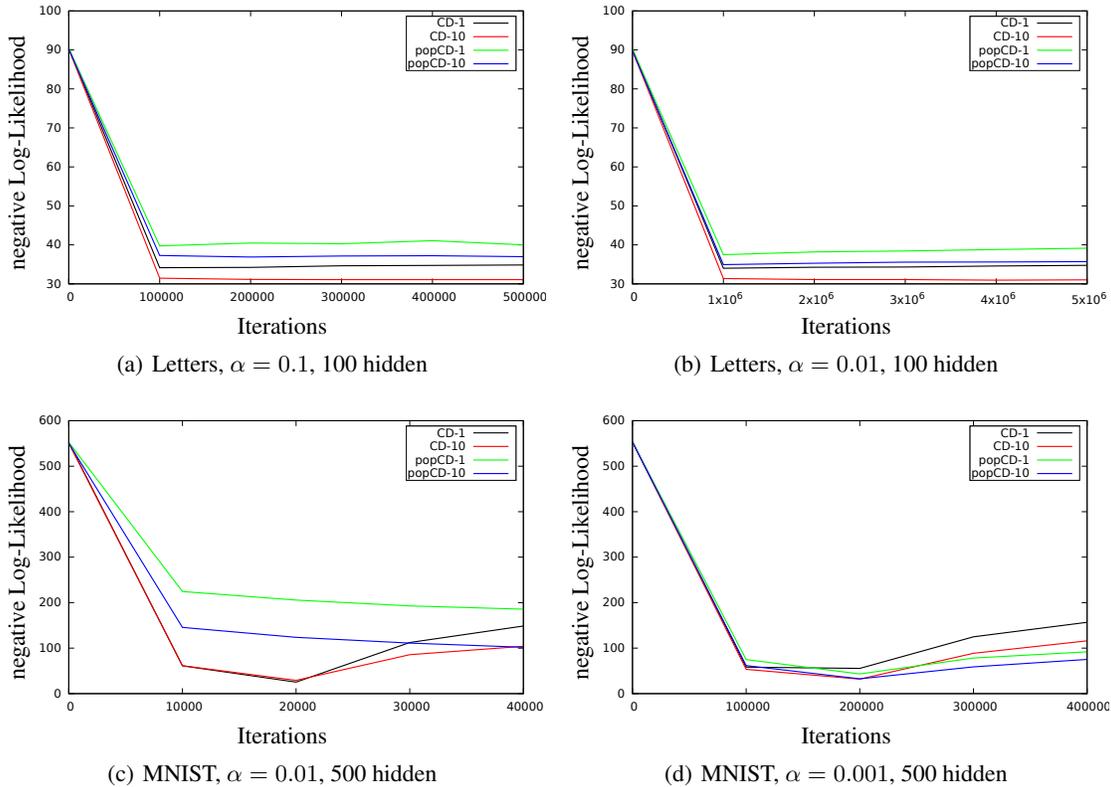
\begin{figure*}[t!]
	\hspace{-0.5cm}\subfigure[Letters, $\alpha=0.1$, 100 hidden]{\label{Letter10001}
		\begin{tikzpicture}
			\draw (0, 0) node[inner sep=0] {\includegraphics[width=0.5\textwidth]{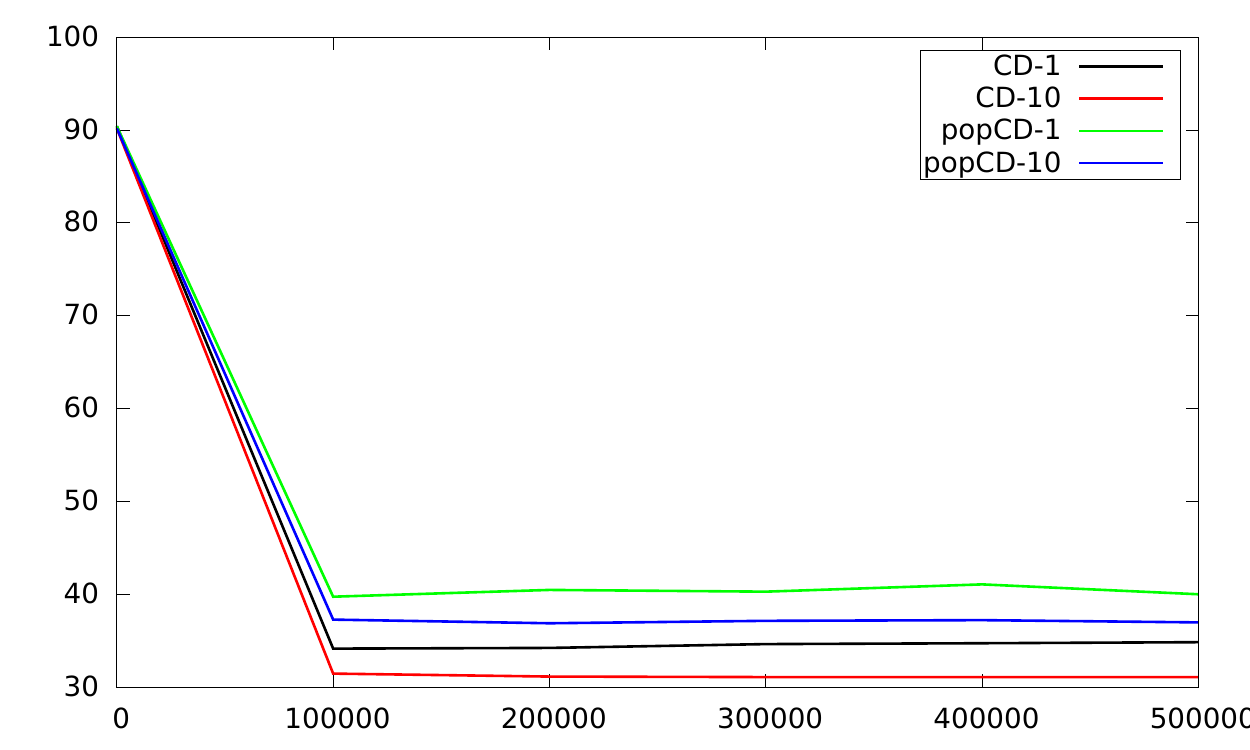}};
			\draw (-3.5, 0) node {\rotatebox{90}{\small negative Log-Likelihood}};
			\draw (0.0, -2.3) node {\small Iterations};
		\end{tikzpicture}
	}
	\subfigure[Letters, $\alpha=0.01$, 100 hidden]{\label{Letter100001}
		\begin{tikzpicture}
			\draw (0, 0) node[inner sep=0] {\includegraphics[width=0.5\textwidth]{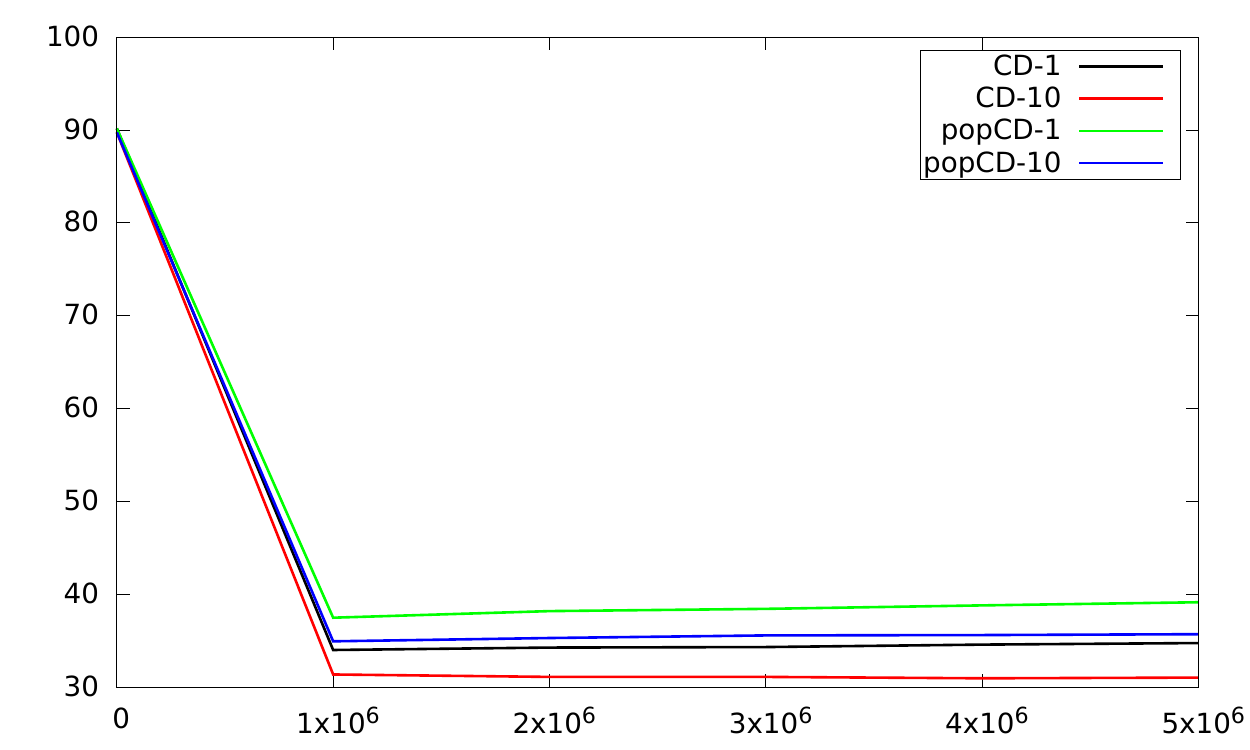}};
			\draw (-3.5, 0) node {\rotatebox{90}{\small negative Log-Likelihood}};
			\draw (0.0, -2.3) node {\small Iterations};
		\end{tikzpicture}	
	}\\
	\hspace*{-0.5cm}\subfigure[MNIST, $\alpha=0.01$, 500 hidden]{\label{MNIST500001}
		\begin{tikzpicture}
			\draw (0, 0) node[inner sep=0] {\includegraphics[width=0.5\textwidth]{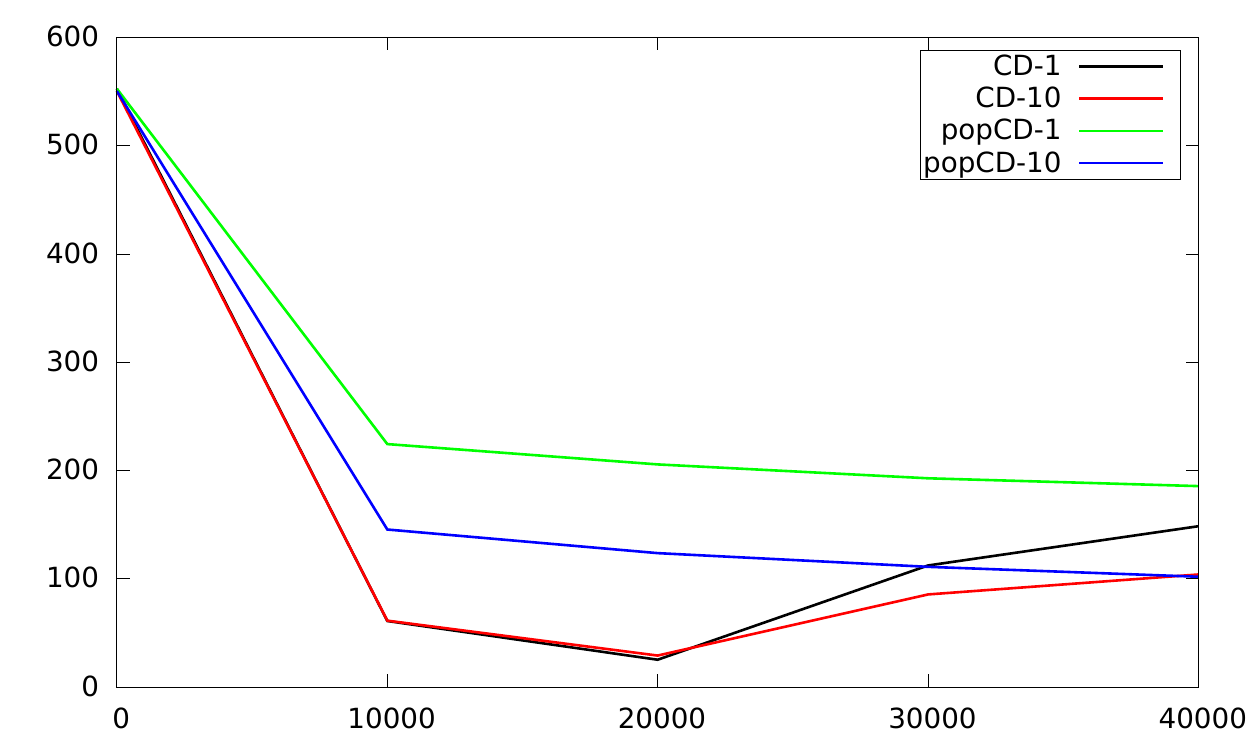}};
			\draw (-3.5, 0) node {\rotatebox{90}{\small negative Log-Likelihood}};
			\draw (0.0, -2.3) node {\small Iterations};
		\end{tikzpicture}
	}
	\subfigure[MNIST, $\alpha=0.001$, 500 hidden]{\label{MNIST5000001}
		\begin{tikzpicture}
			\draw (0, 0) node[inner sep=0] {\includegraphics[width=0.5\textwidth]{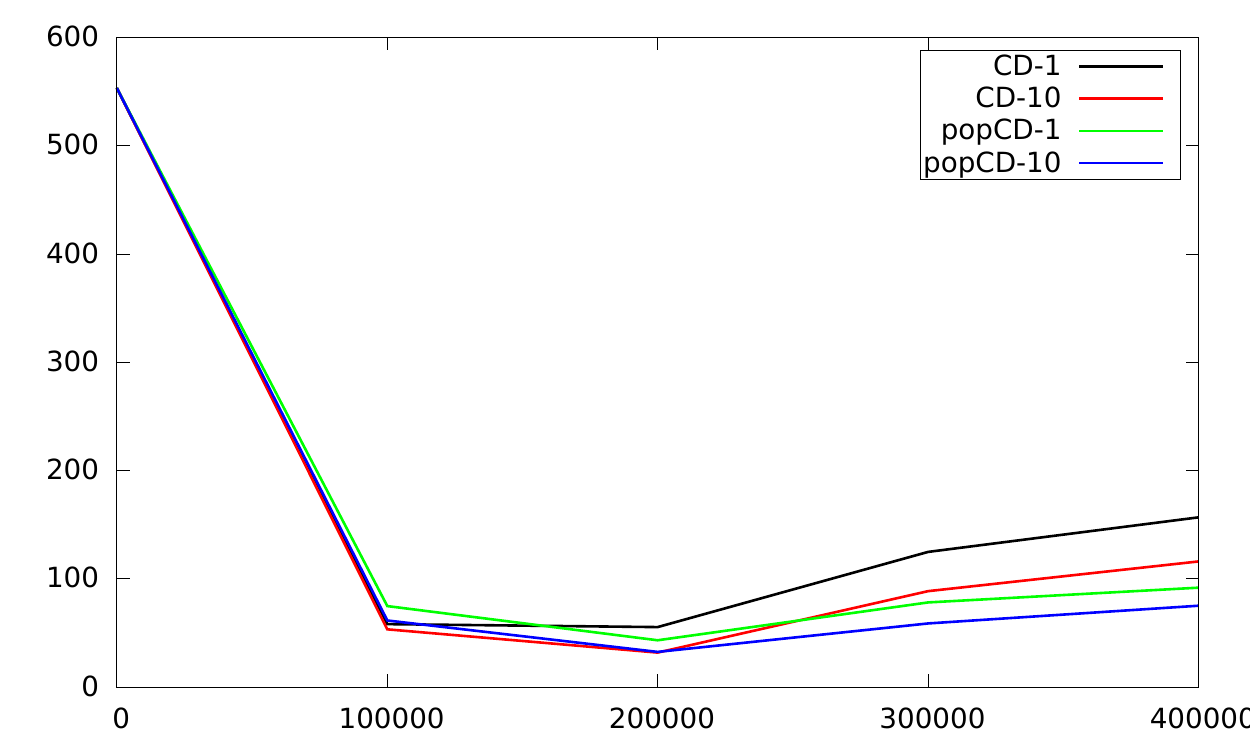}};
			\draw (-3.5, 0) node {\rotatebox{90}{\small negative Log-Likelihood}};
			\draw (0.0, -2.3) node {\small Iterations};
		\end{tikzpicture}	
	}
        \caption{Training curves for MNIST and Letters using larger RBMs.}
	\label{BigResults}
\end{figure*}

\begin{table*}[t!]
      \caption{Estimated Bias and Variance of CD-$k$ and pop-CD-$k$ divided by the number of parameters.}
	\label{VarBiasTable}
\begin{center}
	\begin{tabular}{l| c c | c c | c c}
	 &   & & \multicolumn{2}{|c|}{CD-$k$} &  \multicolumn{2}{|c}{ Pop-CD-$k$} \\
	Problem  & $k$  & hidden & Variance & Bias &  Variance & Bias \\
	\hline
	\input{BaS16}
	\hline
	\input{DM16}
	\hline
	\input{Letters16}
	\hline
	\input{Letters100}
	\hline
	\input{MNIST16}
	\hline
	\input{MNIST500}
	\end{tabular}
      \end{center}
\end{table*}
We compared the new pop-CD-$k$ algorithm experimentally to previously 
proposed methods. We implemented the algorithms in the machine
learning library \textsc{Shark} \citep{igel:08b} and will make the 
code of the experiments available in the case of acceptance of
the manuscript. 

\subsection{Experimental Setup}
We chose two small artificial datasets, \BaS~\citep{mackay2002information} and 
the data set used by~\cite{DesjardinsTempered} and \cite{bruegge:13}
which we will refer to as Artificial Modes, as 
well as two bigger real world datasets,  MNIST~\citep{lecun1998gradient} 
and Letters\footnote{http://ai.stanford.edu/$\sim$btaskar/ocr/}. 
Artificial Modes was created to generate distributions for which
Gibbs-sampling has problems to mix by setting is control parameter to $p=0.001$.
The datasets \BaS{} and Artificial Modes are known to lead to a bias causing
the log-likelihood to diverge for CD and PCD \citep{FischerCDDivergence,DesjardinsTempered}. 
This makes them good benchmarks for analysing the bias of different
gradient estimation strategies.
\par In all experiments, except the ones on \BaS, we performed mini batch
learning. We \emph{a priori} set the size of the batches, and thus also the number of samples generated 
for the estimation of the model mean, to be a divisor of the full dataset size. 
The batch sizes were 32 for \BaS, 500 for Artificial Modes, 246 for Letters, and 500 for MNIST.
We did not use a momentum term in all our experiments.
\par The first set of experiments considered the behaviour of the algorithm in settings
in which the normalisation constant of the RBM can be computed analytically. We selected 
the number of hidden neurons to be 16 and we compared the algorithms 
CD-1, CD-10, PCD-1, pop-CD-1, pop-CD-10, and PT with 10 parallel chains (and inverse temperatures 
uniformly distributed in $[0,1]$). We choose the learning rates
to be $\alpha=0.1$ and $\alpha=0.01$ for all datasets. We let the algorithm run for 50000 and 
10000 iterations (where one iteration corresponds to one step of gradient ascent) on 
\BaS{} and Artificial Modes, respectively.
On MNIST we trained for 5000 iterations with $\alpha=0.1$ and for 20000 with $\alpha=0.01$, 
and on Letters we trained for 10000 iterations with $\alpha=0.1$ and 30000 iterations with 
$\alpha=0.01$. We evaluated the log-likelihood every 100 iterations. All experiments were 
repeated 25 times and the reported log-likelihood values are  the means over all trials.
\par In our second set of experiments, we trained RBMs with a large number of hidden neurons. 
Here the normalisation constant was estimated using Annealed Importance Sampling (AIS, \citealp{NealAIS})
with 512 samples and 50000 intermediate distributions.
The learning parameters  were $\alpha=0.1$ with $5 \cdot 10^5$ iterations and $\alpha=0.01$ 
with $5 \cdot 10^6$ iterations for Letters and for MNIST $\alpha=0.01$ with $4 \cdot 10^4$ iterations and $\alpha=0.001$ 
with $4 \cdot 10^5$ iterations. Due to the long running times we do not
consider PCD and PT and only show one trial for MNIST. For Letters,
the results are the average over 10 trials.
\par In a third set of experiments, we investigated the bias and the
variance of the methods. We first trained an RBM using CD-1
with $\alpha=0.1$ and 16 hidden neurons on all four problems.
The number of training iterations was 10000 for \BaS, Artificial Modes 
and Letters while 5000 on MNIST. 
Furthermore, we repeated the large experiments
by training an RBM with $\alpha=0.01$ and 500 hidden units for 15000 iterations on MNIST and 
with $\alpha=0.1$, 100 hidden units and $10^5$ iterations on Letters.
After that, we  calculated a ``ground truth `` estimate of the true gradient.
This was done by PT with 100 intermediate chains. The chain was given 50000 iterations burn-in time 
and afterwards 200000 samples were taken to compute the estimate.
Then, we generated 50000 gradient estimates of CD-$k$ and pop-CD-$k$ for $k \in \{1,10\}$. 
Finally, we estimated the bias of the methods as the squared distance between the empirical average 
of the methods and the estimated ground truth
and the variance as the expectation of the squared distance between the single samples and their mean.

\subsection{Results}
\par The results of the first set of experiments can be seen in Figure~\ref{SmallResults1} 
for the artificial datasets and in Figure~\ref{SmallResults2} for MNIST and Letters. 
Figure~\ref{SmallResults1} shows that pop-CD-1 as well as pop-CD-10 
outperformed CD-1 and CD-10. The proposed algorithm
reached significantly higher log-likelihood values
and, while CD and PCD diverged, pop-CD did not show any sign of divergence.
Using $k=1$ and $k=10$ sampling steps lead to almost the same
performance of pop-CD.
The efficient pop-CD-1 performed on a par or better than the
computationally more expensive PT-10.
On the two real datasets CD did not cause any divergence. 
However, while all algorithms performed almost the same on Letters, 
pop-CD-1 still reaches higher likelihood values than CD-1 on MNIST,
which stopped showing improvement after 3000 iterations in Figure~\ref{MNIST1601}.
Furthermore, one can observe that pop-CD performed slightly
better with a smaller learning rate of  $\alpha=0.01$
than with a learning rate of $\alpha=0.1$.

\par The results for the second set of experiments are given in Figure~\ref{BigResults}. 
The plots for the Letters data set (Figures \ref{Letter10001} and
\ref{Letter100001}) look qualitatively the same, aside from the scaling of the abscissa. In both cases,
CD-1 and CD-10 performed better than pop-CD-1 and pop-CD-10. 
On MNIST (Figures \ref{MNIST500001} and \ref{MNIST5000001}), both CD-1 and CD-10 diverged after
the initial learning phase. For $\alpha=0.01$ pop-CD showed a very slow learning progress, while for
$\alpha=0.001$ all four algorithms exhibit similar behaviour. While all algorithms diverge in the long
run, divergence was less pronounced for pop-CD-$k$ than CD-$k$.

Table \ref{VarBiasTable} shows the measured bias and variance for
the third experimental setup.  For the experiments
with 16 hidden neurons,  CD-$k$ always had a smaller variance than pop-CD-$k$, 
often by an order of magnitude or more. The bias of pop-CD-$k$ was in contrast much smaller,
which explains the experimental results we observed. In the
experiments with 500  hidden units,
pop-CD-$k$ had a larger bias as well.

\section{DISCUSSION}
\par In most experiments, the new method performed on a par 
or even better compared to computational much more expensive sampling 
techniques such as PT. However, for problems where the bias of CD does not 
notably impair the learning,  pop-CD can be slower than CD,
because it may require smaller learning rates for stochastic gradient descent. 
\par The fact that pop-CD fares better with smaller learning rates 
can be explained by a larger variance, as measured in all our
experiments, see Table~\ref{VarBiasTable}.
This is due to the bias-variance-trade-off, which is
a well known issue for importance sampling based estimators.
While having {potentially} low bias, the variance of importance sampling based estimators 
depends on how well the proposal function approximates the target distribution.
Thus, our results indicate that the used proposal function can lead to a high variance and may 
thus not be the optimal choice for complex target distributions.
When working with a small sample size compared to the difficulty of
the estimation problem or with a bad proposal distribution, the estimate of the normalisation constant can
have a large variance, which leads to a biased estimate of the inverse
normalisation constant. This is supported by our results in the
experiments with a large number of hidden units  as well as by the bias estimates, see Table~\ref{VarBiasTable}.

\par The observation that more steps of Gibbs-sampling help when using pop-CD for 
larger models can be explained by the analysis in Section \ref{Seq:AnalysWeights}.
The variance of the estimator depends on how well the sampled $\h'$---and, thus, the proposal distribution
$p(\vis|\h')$---approximates the RBM distribution. 
The higher $k$ the closer $q_i^{(k-1)}(\h)$ gets to $p(\h)$. This reduces the variance of the weights
and affects the estimator of the gradient as well as the estimator of the normalisation constant.
This is supported by the results in Table~\ref{VarBiasTable} for
the different values of $k$. 

\section{CONCLUSIONS}
Improving the sample 
quality of the Markov chain (by increasing the number of sampling steps or 
employing advanced sampling techniques like PT) is not the only direction 
leading to low bias estimates---importance sampling can also reduce the
bias efficiently.

We have proposed a new variant of the Contrastive Divergence (CD)
algorithm inspired by the Population Monte Carlo method. 
The new learning algorithm, termed Population-CD (pop-CD),
incorporates importance weights into the sampling scheme of CD. 
This makes the bias of the log-likelihood gradient estimate independent of the 
Markov chain and sampling operators used, with negligible  
computational overhead compared to CD. However, this comes at
the cost of an increased variance of the estimate. Further the bias of the method
depends on how well the empirical mean of the importance weights estimates
the normalisation constant of the distribution.

\par In contrast to CD, pop-CD is consistent. For problems with a
small number of hidden neurons it clearly outperforms CD with the same
computational cost, leading to higher likelihood values.  However, for
{RBMs with many hidden neurons}, pop-CD may require a large number of samples to achieve
a smaller bias than CD---therefore, CD is still recommended in that case.  
The reason for this  is that our  current estimator relies on a set of simple proposal distributions, $p(\vis|\h)$, which
do not use information about  the distribution of the samples
$q(\h')$. Thus, the estimator
can be seen as replacing the unknown distribution of $q(\h')$ by a known distribution, which is
worse than the distribution of the samples that is available through
the Markov Chain. Hence, future work must investigate whether there exists 
a better choice for the proposal distribution
bridging the gap between $q(\h')$ and the true distribution $p(\h)$.

\subsubsection*{Acknowledgements}
Christian and Oswin  acknowledge support from the Danish National Advanced Technology Foundation through project 
``Personalized breast cancer screening''.

\vfill
\pagebreak
\appendix
\section{Pop-CD Algorithm}

\begin{algorithm}[hb!]
\label{alg:pop-CD}
\caption{$k$-step pop-CD} 
\KwIn{RBM $(V_1,\dots , V_m,H_1,\dots , H_n)$, training batch $S$ of size $\ell$}
\KwOut{gradient approximation $\Delta w_{ij}$, $\Delta b_j$ and
  $\Delta c_i$ for $i=1,\dots , n$, $j=1,\dots , m$}
init $\Delta w_{ij}^+=\Delta w_{ij}^-=\Delta b_j^+=\Delta b_j^-=\Delta c_i^+=\Delta c_i^-=0$ 
for $i=1,\dots , n$, $j=1,\dots , m$, $\omega_S= 0$ \\
\ForAll{$\vec v\in S$} 
{
    $\vec v^{(0)} \leftarrow \vec v$ \\
    \For{$t=0,\dots ,k-1$}
    {
	sample $\h^{(t)} \sim p(\h\,|\,\vis^{(t)})$\\
	sample $\vis^{(t+1)} \sim p(\vis\,|\,\h^{(t)})$\label{PVHAlgorithm}\\
    }
    $\omega \leftarrow \frac{p(\vis^{(k)})}{p(\vis^{(k)}|\h^{(k-1)})}$\label{OmegaAlgorithm}\\
    $\omega_S \leftarrow \omega_S+\omega$\\
    \For{$i=1,\dots ,n$,  $j=1,\dots ,m$}
    {  
	  $  \Delta w_{ij}^+  \leftarrow  \Delta w_{ij}^+ + p({H_i} = 1 \,|\, \vis^{(0)}) \cdotp v^{(0)}_j $\\
	  $  \Delta w_{ij}^-  \leftarrow  \Delta w_{ij}^- + \omega p({H_i} = 1 \,|\, \vis^{(k)})\cdotp v^{(k)}_j$\\
    }
    \For{$j=1,\dots ,m$}
    {
    $  \Delta b_j^+  \leftarrow  \Delta b_j^+ + v^{(0)}_j$ \\
    $  \Delta b_j^-  \leftarrow  \Delta b_j^- + \omega v^{(k)}_j $ 
    }
    \For{$i=1,\dots ,n$}
    {
     $ \Delta c_i^+  \leftarrow  \Delta c_i^+  +  p({H_i} = 1 \,|\, \vec v^{(0)}) $
     $ \Delta c_i^-  \leftarrow  \Delta c_i^- +   \omega p({H_i} = 1 \,|\, \vec v^{(k)})$
    }
}
\For{$i=1,\dots ,n$, $j=1,\dots ,m$}
{
   $\Delta w_{ij}  \leftarrow  \frac 1 {\ell} \Delta w_{ij}^+ - \frac 1 {\omega_S} \Delta w_{ij}^-$ \\
   $\Delta b_{j}  \leftarrow   \frac 1 {\ell} \Delta b_{j}^+ - \frac 1 {\omega_S} \Delta b_{j}^-$ \\
   $\Delta c_{i}  \leftarrow   \frac 1 {\ell} \Delta c_{i}^+ - \frac 1 {\omega_S} \Delta c_{i}^-$ \\
} 
\end{algorithm} 

\end{document}

%% file: BaS16.tex
\multirow{2}{*}{\BaS}& 1 & 16 & 5.34874e-05 & 0.0242761 & 0.000527033 & 0.000156001\\
& 10 & 16 & 0.000238678 & 0.0233395 & 0.0011887 & 0.000139806\\

%% file: DM16.tex
\multirow{2}{*}{Artificial Modes}& 1 & 16 & 2.3712e-06 & 0.177744 & 0.000372144 & 7.99706e-09\\
& 10 & 16 & 5.07146e-06 & 0.176977 & 0.000372545 & 6.29318e-09\\

%% file: Letters16.tex
\multirow{2}{*}{Letters}& 1 & 16 & 0.000332673 & 0.00213099 & 0.0168316 & 0.000681575\\
& 10 & 16 & 0.000434146 & 0.0017085 & 0.0148595 & 0.000300025\\

%% file: Letters100.tex
\multirow{2}{*}{Letters}& 1 & 100& 7.72345e-05 & 0.00321983 & 0.0115259 & 
0.0133755\\
& 10 & 100 & 0.000186199 & 0.00297003 & 0.0128639 & 0.00942309\\

%% file: MNIST16.tex
\multirow{2}{*}{MNIST}& 1 & 16 & 0.000143567 & 0.0145665 & 0.00377791 & 0.000746869\\
& 10 & 16 & 0.000156719 & 0.0136011 & 0.00244656 & 0.000694042\\

%% file: MNIST500.tex
\multirow{2}{*}{MNIST}& 1 & 500 & 2.98891e-05 & 0.00756879 & 0.00581998 & 0.00974112\\
& 10 & 500 & 6.47712e-05 & 0.00670447 & 0.0059567 & 0.00289032\\

%% file: KrauseFischerIgelPopCDICLR-arXiv.bbl
\begin{thebibliography}{16}
\providecommand{\natexlab}[1]{#1}
\providecommand{\url}[1]{\texttt{#1}}
\expandafter\ifx\csname urlstyle\endcsname\relax
  \providecommand{\doi}[1]{doi: #1}\else
  \providecommand{\doi}{doi: \begingroup \urlstyle{rm}\Url}\fi

\bibitem[Br{\"u}gge et~al.(2013)Br{\"u}gge, Fischer, and Igel]{bruegge:13}
Br{\"u}gge, K, Fischer, A., and Igel, C.
\newblock The flip-the-state transition operator for restricted {B}oltzmann
  machines.
\newblock \emph{Machine Learning}, 13:\penalty0 53--69, 2013.

\bibitem[Capp{\'e} et~al.(2004)Capp{\'e}, Guillin, Marin, and
  Robert]{cappe2004population}
Capp{\'e}, O., Guillin, A., Marin, J.-M., and Robert, C.~P.
\newblock Population {Monte Carlo}.
\newblock \emph{Journal of Computational and Graphical Statistics}, 13\penalty0
  (4), 2004.

\bibitem[Desjardins et~al.(2010)Desjardins, Courville, Bengio, Vincent, and
  Delalleau]{DesjardinsTempered}
Desjardins, G., Courville, A.~C., Bengio, Y., Vincent, P., and Delalleau, O.
\newblock Tempered {Markov chain Monte Carlo for training of restricted
  Boltzmann} machines.
\newblock In \emph{Proceedings of the 13th International Conference on
  Artificial Intelligence and Statistics (AISTATS 2010)}, volume~9 of
  \emph{JMLR: W\&C}, pp.\  145--152, 2010.

\bibitem[Fischer \& Igel(2010)Fischer and Igel]{FischerCDDivergence}
Fischer, A. and Igel, C.
\newblock Empirical analysis of the divergence of {Gibbs} sampling based
  learning algorithms for {Restricted Boltzmann Machines}.
\newblock In Diamantaras, K., Duch, W., and Iliadis, L.~S. (eds.),
  \emph{International Conference on Artificial Neural Networks (ICANN 2010)},
  volume 6354 of \emph{LNCS}, pp.\  208--217. Springer-Verlag, 2010.

\bibitem[Fischer \& Igel(2011)Fischer and Igel]{FischerCDBias}
Fischer, A. and Igel, C.
\newblock Bounding the bias of contrastive divergence learning.
\newblock \emph{Neural Computation}, 23\penalty0 (3):\penalty0 664--673, 2011.

\bibitem[Fischer \& Igel(2014)Fischer and Igel]{fischer:13}
Fischer, A. and Igel, C.
\newblock Training restricted {B}oltzmann machines: {An} introduction.
\newblock \emph{Pattern Recognition}, 47:\penalty0 25--39, 2014.

\bibitem[Hinton(2002)]{HintonCD}
Hinton, G.~E.
\newblock Training products of experts by minimizing contrastive divergence.
\newblock \emph{Neural Computation}, 14\penalty0 (8):\penalty0 1771--1800,
  2002.

\bibitem[Igel et~al.(2008)Igel, Glasmachers, and Heidrich-Meisner]{igel:08b}
Igel, C., Glasmachers, T., and Heidrich-Meisner, V.
\newblock Shark.
\newblock \emph{Journal of Machine Learning Research}, 9:\penalty0 993--996,
  2008.

\bibitem[LeCun et~al.(1998)LeCun, Bottou, Bengio, and
  Haffner]{lecun1998gradient}
LeCun, Y., Bottou, L., Bengio, Y., and Haffner, P.
\newblock Gradient-based learning applied to document recognition.
\newblock \emph{Proceedings of the IEEE}, 86\penalty0 (11):\penalty0
  2278--2324, 1998.

\bibitem[MacKay(2002)]{mackay2002information}
MacKay, D. J.~C.
\newblock \emph{Information theory, inference, and learning algorithms},
  volume~7.
\newblock Cambridge University Press, 2002.

\bibitem[Neal(2001)]{NealAIS}
Neal, R.~M.
\newblock Annealed importance sampling.
\newblock \emph{Statistics and Computing}, 11\penalty0 (2):\penalty0 125--139,
  2001.

\bibitem[Salakhutdinov(2009)]{SalakhutdinovTemperedTransitions}
Salakhutdinov, R.
\newblock Learning in {Markov} random fields using tempered transitions.
\newblock In Bengio, Y., Schuurmans, D., Lafferty, J., Williams, C. K.~I., and
  Culotta, A. (eds.), \emph{Advances in Neural Information Processing Systems
  22 (NIPS)}, pp.\  1598--1606, 2009.

\bibitem[Schulz et~al.(2010)Schulz, M{\"u}ller, and
  Behnke]{schulz2010investigating}
Schulz, H., M{\"u}ller, A., and Behnke, S.
\newblock Investigating convergence of restricted {B}oltzmann machine learning.
\newblock In Lee, H., Ranzato, M., Bengio, Y., Hinton, G.~E, LeCun, Y., and Ng,
  A.~Y. (eds.), \emph{NIPS 2010 Workshop on Deep Learning and Unsupervised
  Feature Learning}, 2010.

\bibitem[Smolensky(1986)]{smolensky}
Smolensky, P.
\newblock Information processing in dynamical systems: Foundations of harmony
  theory.
\newblock In Rumelhart, D.~E. and McClelland, J.~L. (eds.), \emph{Parallel
  Distributed Processing: Explorations in the Microstructure of Cognition, vol.
  1: Foundations}, pp.\  194--281. MIT Press, 1986.

\bibitem[Sutskever \& Tieleman(2010)Sutskever and
  Tieleman]{sutskever2010convergence}
Sutskever, I. and Tieleman, T.
\newblock On the convergence properties of contrastive divergence.
\newblock In \emph{Proceedings of the 13th International Conference on
  Artificial Intelligence and Statistics (AISTATS 2010)}, volume~9 of
  \emph{JMLR: W\&C}, pp.\  789--795, 2010.

\bibitem[Tieleman(2008)]{tieleman2008training}
Tieleman, T.
\newblock Training restricted {B}oltzmann machines using approximations to the
  likelihood gradient.
\newblock In \emph{Proceedings of the 25th International Conference on Machine
  Learning (ICML 2008)}, pp.\  1064--1071. ACM, 2008.

\end{thebibliography}
